%% file: main.tex
\documentclass[conference]{IEEEtran}
\IEEEoverridecommandlockouts

\usepackage{cite}
\usepackage{amsmath,amssymb,amsfonts}
\usepackage{graphicx}
\usepackage{textcomp}
\usepackage{xcolor}

\input{section/preamble}

\begin{document}

\title{Matrix Profile for Anomaly Detection on Multidimensional Time Series}

\input{section/author}

\maketitle

\input{section/abstract}
\input{section/introduction}
\input{section/related}
\input{section/background}

\input{section/method}
\input{section/experiment}

\input{section/conclusion}

\balance
\bibliographystyle{IEEEtran}
\bibliography{section/reference}

\end{document}

%% file: section/preamble.tex
\usepackage{graphicx}
\usepackage{stfloats}
\usepackage{wrapfig}
\usepackage{lipsum}  
\usepackage{url} 

\newlength\savewidth
\newcommand\shline{\noalign{\global\savewidth\arrayrulewidth
                            \global\arrayrulewidth 1.5pt}%
                   \hline
                   \noalign{\global\arrayrulewidth\savewidth}
                   }

\usepackage{xcolor}

\usepackage{algorithm}
\usepackage[noend]{algpseudocode}

\newcommand*\Input[1]{\Statex \textbf{Input:} #1}
\newcommand*\Output[1]{\Statex \textbf{Output:} #1}
\algrenewcommand\alglinenumber[1]{#1}

\usepackage{multirow}

\usepackage{enumitem}
\setlist[itemize]{leftmargin=*, topsep=.0em, itemsep=0pt, parsep=0pt, partopsep=0pt}
\setlist[enumerate]{leftmargin=*, topsep=.0em, itemsep=0pt, parsep=0pt, partopsep=0pt}

\usepackage{amsthm}
\newtheoremstyle{def_style}
  {0.5em}      
  {0.5em}      
  {}          
  {}          
  {\bfseries} 
  {.}         
  {0.5em}      
  {}          

\theoremstyle{def_style}
\newtheorem{define}{Definition}

\theoremstyle{def_style}
\newtheorem{prob}{Problem}

\usepackage{amsmath}
\usepackage{bbm}

\usepackage{collectbox}



\usepackage{mathtools}

\usepackage{balance}

%% file: section/author.tex

\author{
\IEEEauthorblockN{Chin-Chia Michael Yeh\IEEEauthorrefmark{1}, Audrey Der\IEEEauthorrefmark{2}, Uday Singh Saini\IEEEauthorrefmark{1}, Vivian Lai\IEEEauthorrefmark{1}, Yan Zheng\IEEEauthorrefmark{1},\\Junpeng Wang\IEEEauthorrefmark{1}, Xin Dai\IEEEauthorrefmark{1}, Zhongfang Zhuang\IEEEauthorrefmark{1}, Yujie Fan\IEEEauthorrefmark{1}, Huiyuan Chen\IEEEauthorrefmark{1},\\Prince Osei Aboagye\IEEEauthorrefmark{1}, Liang Wang\IEEEauthorrefmark{1}, Wei Zhang\IEEEauthorrefmark{1}, Eamonn Keogh\IEEEauthorrefmark{2}}
\IEEEauthorblockA{\IEEEauthorrefmark{1}\textit{Visa Research}, \IEEEauthorrefmark{2}\textit{UC Riverside}\\
miyeh@visa.com}
}

%% file: section/abstract.tex
\begin{abstract}
The Matrix Profile (MP), a versatile tool for time series data mining, has been shown effective in time series anomaly detection (TSAD). 
This paper delves into the problem of anomaly detection in multidimensional time series, a common occurrence in real-world applications. 
For instance, in a manufacturing factory, multiple sensors installed across the site collect time-varying data for analysis. 
The Matrix Profile, named for its role in profiling the matrix storing pairwise distance between subsequences of univariate time series, becomes complex in multidimensional scenarios. 
If the input univariate time series has $n$ subsequences, the pairwise distance matrix is a $n \times n$ matrix. 
In a multidimensional time series with $d$ dimensions, the pairwise distance information must be stored in a $n \times n \times d$ tensor. 
In this paper, we first analyze different strategies for condensing this tensor into a profile vector. 
We then investigate the potential of extending the MP to efficiently find $k$-nearest neighbors for anomaly detection. 
Finally, we benchmark the multidimensional MP against 19 baseline methods on 119 multidimensional TSAD datasets. 
The experiments covers three learning setups: unsupervised, supervised, and semi-supervised. 
MP is the only method that consistently delivers high performance across all setups.

To ensure complete transparency and facilitate future research, our full Matrix Profile-based implementation, which includes newly added evaluations against the TSB-AD benchmark, is publicly available at: \url{https://github.com/mcyeh/mmpad_tsb}
\end{abstract}


\begin{IEEEkeywords}
time series, multidimensionality, matrix profile, discord mining, anomaly detection
\end{IEEEkeywords}

%% file: section/introduction.tex
\section{Introduction}
\label{sec:intro}
The Matrix Profile (MP) is an effective and efficient tool for detecting anomalous patterns within a time series~\cite{yeh2016matrix,lu2022matrix}. 
It detects anomalies using the concept of time series discord~\cite{keogh2005hot,yeh2018towards,yeh2023sketching}, where the level of anomalousness of a subsequence is measured using the nearest neighbor distance (e.g., $z$-normalized Euclidean distance). 
In other words, if a subsequence and its most similar counterpart within the same time series are  distant in Euclidean space, then the subsequence is likely an anomaly.

The primary challenge in extending MP to multidimensional time series for anomaly detection is that the anomalous pattern usually spans only a handful of dimensions in the multidimensional time series~\cite{tafazoli2023matrix}. 
As a result, we cannot simply sum the distances computed using each individual dimension, as this would bury the anomalous pattern under all the dimensions consisting of normal patterns. 
This challenge is referred to as the \textit{K of N anomaly detection problem} in~\cite{tafazoli2023matrix}. 
Fig.~\ref{fig:kn_problem} demonstrates that naively computing MP using all dimensions does not solve the $K$ of $N$ anomaly detection problem.
In the figure, we have an eight-dimensional time series, of which only one dimension consists of the anomalous pattern. 

\input{insert/fig_kn_problem}

If we compute MP using all dimensions, the result would be the Matrix Profile (naive) depicted in the figure. 
Unfortunately, this fails to correctly identify the location of the anomalous pattern. 
The maximum value, marked with an orange X in the figure, indicates the location of the subsequence with the largest nearest neighbor distance, but it is not temporally close to the anomalous pattern. 
This observation suggested that it is more effective to compute MP using only a subset of the dimensions rather than all of them. 
However, identifying the correct combination of dimensions using a brute force solution has a time complexity of $O(d!)$, where $d$ represents the number of dimensions in the time series. 
To resolve this issue more efficiently, we employ a greedy algorithm where the dimensions are selected in order from most to least anomalous. 
This dimension selection process can be implemented using any sorting algorithm once the distances between subsequences have been computed, thereby reducing the time complexity from $O(d!)$ to $O(d \log{d})$.

Recall that MP is a vector summarizing the pairwise distances between each pair of subsequences in a time series. 
These distances can be stored in a matrix where the $(i, j)$ position stores the distance between the $i$-th and $j$-th subsequences in the time series.
The Matrix Profile gets its name from its function of ``profiling" this pairwise distance matrix by identifying the nearest neighbor for each subsequence. 
Conceptually, the computation process of MP can be divided into two stages: 1) computation of the pairwise distances, and 2) identification of nearest neighbors. 
Given that the computation process of MP involves these two stages and that a sorting algorithm needs to be applied after the distances are computed, this sorting algorithm can be applied either before or after the neighbor identification step. 
We refer to the variant where the sorting is done after identifying the nearest neighbor as post-sorting, and the alternative as pre-sorting.

Returning to the example shown in Fig.~\ref{fig:kn_problem}, if we use the aforementioned variants to properly compute MP for multidimensional time series, we would obtain the other two MPs (i.e., post-sorting and pre-sorting). 
Both variants successfully identify the anomalous pattern at the correct time location, as the maximum value in both MPs occurs around the anomalous moment. 
In Section~\ref{sec:method}, we discuss the strengths and weaknesses of both variants in more detail, and provide recommendations on choosing the best variant for the anomaly detection problems readers may encounter.

In addition to addressing the problem associated with the computation of the multidimensional MP, we also discuss how to effectively utilize MP for anomaly detection. 
Specifically, we explain how to apply MP under various anomaly detection learning setups, namely unsupervised, supervised, and semi-supervised. 
We extend MP to efficiently find the $k$-nearest neighbors, rather than just the nearest neighbor, to mitigate the ``twin freak" problem~\cite{lu2022matrix}, which emerges when an anomalous pattern recurs.
The proposed $k$-nearest neighbors finding algorithm is an order of magnitude faster than the alternative solutions.
Utilizing the presented knowledge, we evaluated the multidimensional MP on 119 multidimensional time series anomaly detection datasets from the TimeEval package~\cite{schmidl2022anomaly}. 
Compared to the other 19 baseline methods, the multidimensional MP is the only method that consistently achieves high averaged performance across all three learning setups.

Finally, alongside this paper, we provide a comprehensive open-source implementation of our methodology \cite{yeh2026mmpad}. Within this repository, we have further benchmarked our approach against the rigorous TSB-AD framework \cite{liu2024elephant, yeh2026matrix} to establish new, reproducible baselines for the community.

%% file: insert/fig_kn_problem.tex
\begin{figure}[htp]
\centerline{
\includegraphics[width=0.99\linewidth]{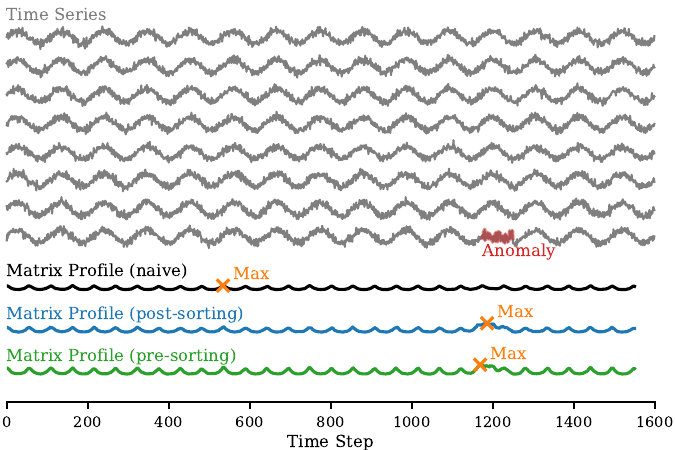}
}
\caption{
Extending the Matrix Profile to multi-dimensional time series is a non-trivial task, given that anomalous patterns are likely to appear in only a small set of dimensions.
}
\label{fig:kn_problem}
\end{figure}

%% file: section/related.tex
\section{Related Work}
Anomaly detection algorithms can be categorized into six families: 1) forecasting, 2) reconstruction, 3) distance, 4) encoding, 5) distribution, and 6) tree methods~\cite{schmidl2022anomaly}.

Forecasting-based methods such as Torsk~\cite{heim2019adaptive}, Telemanom~\cite{hundman2018detecting}, and DeepAnT~\cite{munir2018deepant} detect anomalies within time series by first building a forecasting system and then measuring the difference between the predicted and actual time series. 
Reconstruction-based methods like OmniAnomaly~\cite{su2019robust} and RobustPCA~\cite{paffenroth2018robust} project the input time series subsequence to a low-dimensional representation, then use a decoder to reconstruct the subsequence. 
The anomaly of the subsequence is measured by the difference between the original and reconstructed subsequences.

Distance-based methods such as Matrix Profile (MP)~\cite{yeh2016matrix}, $k$-Means~\cite{yairi2001fault}, $k$NN~\cite{ramaswamy2000efficient}, CBLOF~\cite{he2003discovering}, LOF~\cite{breunig2000lof}, PCC~\cite{shyu2003novel}, HBOS~\cite{goldstein2012histogram}, COF~\cite{tang2002enhancing}, and Hybrid $k$NN~\cite{song2017hybrid} compute the anomaly score by measuring the distance between different subsequences. 
Encoding-based methods like MultiHMM~\cite{li2017multivariate} encode the input time series subsequence to a low-dimensional representation and compute the anomaly score in the low-dimensional space.
Distribution-based methods like COPOD~\cite{li2020copod} fit a distribution model to the data and estimate the probability of the input data being anomalous. 
Tree-based methods such as EIF~\cite{hariri2019extended}, iForest~\cite{liu2008isolation}, IF-LOF~\cite{cheng2019outlier}, and HIF~\cite{marteau2017hybrid} utilize an ensemble of random trees to partition the data and compute the anomaly score based on the partition.

The authors of~\cite{schmidl2022anomaly} have conducted a comprehensive evaluation of various time series anomaly detection (TSAD) methods and showcased the great performance of MP on univariate TSAD datasets.
However, they did not evaluate MP on multidimensional time series datasets.
A method which extend MP to multidimensional time series is provided in~\cite{tafazoli2023matrix}, but it does not address certain types of multidimensional anomalies (i.e., correlation anomalies~\cite{baumgartner2023mtads}) as discussed in Section~\ref{sec:multi_mp}.
To our knowledge, our paper is the first to offer a comprehensive study on the application of MP to multidimensional TSAD problems.

%% file: section/background.tex
\section{Background}
\label{sec:background}
In this section, we introduce key definitions that are crucial for understanding the problem formulation and the methodology. 
We then present the three distinct formulations of the time series anomaly detection (TSAD) problem that we have adopted in our study. 
For notation, lowercase letters (e.g., $x$), boldface lowercase letters (e.g., $\mathbf{x}$), uppercase letters (e.g., $X$), and boldface uppercase letters (e.g., $\mathbf{X}$) are used to represent scalars, vectors, matrices, and tensors, respectively. 
We use $\mathbb{B}$ to denote the set of Boolean numbers.

\subsection{Definition}
\label{sec:definition}
First, we establish the definition of a \textit{multidimensional time series}.

\begin{define} 
A multidimensional time series, consisting of $n$ time steps and $d$ dimensions, is stored in a matrix $T \in \mathbb{R}^{n \times d}$. 
Note that we use $T[i:i + m, j]$ to denote a subsequence of length $m$ that begins at the $i$-th time step in the $j$-th dimension. 
\end{define}

As the Matrix Profile summarizes pairwise distances between subsequences, we introduce the concept of a \textit{pairwise distance tensor}. 
This tensor stores the pairwise distances between all subsequences in one time series and all subsequences in another time series.

\begin{define} 
Given a time series $T_1 \in \mathbb{R}^{n_1 \times d}$, another time series $T_2 \in \mathbb{R}^{n_2 \times d}$, and a subsequence length~$m$, the pairwise distance tensor $\mathbf{D} \in \mathbb{R}^{(n_1 - m + 1) \times (n_2 - m + 1) \times d}$ stores the pairwise distances between subsequences in $T_1$ and subsequences in $T_2$. 
Each element within $\mathbf{D}$ is defined as: 
\begin{equation} 
    \mathbf{D}[i, j, k] \gets \textsc{Distance}(T_1[i:i + m, k], T_2[j:j + m, k]) 
\end{equation} 
where the function $\textsc{Distance}(\cdot,\cdot) \rightarrow \mathbb{R}$ computes the distance between the input subsequences. 
\label{def:distance} \end{define}

Following~\cite{yeh2016matrix,yeh2017matrix,yeh2022error}, we use the $z$-normalized Euclidean distance as our distance function due to its efficacy and efficiency. 
Although Definition~\ref{def:distance} is defined for a given pair of time series (AB-join~\cite{yeh2016matrix}), it can also be computed between all the subsequences within a single time series (self-join~\cite{yeh2016matrix}).

\subsection{Problem Formulation}
\label{sec:problem}
There are three formulations (or learning setups) for the multidimensional TSAD problem: unsupervised, supervised, and semi-supervised~\cite{schmidl2022anomaly}. 
The problem statement for the \textit{unsupervised} variant is as follows:

\begin{prob} 
Given a test time series~$T_\text{test} \in \mathbb{R}^{n_\text{test} \times d}$, the objective of \textit{unsupervised anomaly detection} is to output an anomaly score vector~$\hat{\mathbf{y}}_\text{test} \in \mathbb{R}^{n_\text{test}}$ where $\hat{\mathbf{y}}_\text{test}[i]$ stores the score associated with the $i$-th time step. 
\end{prob}

Under the unsupervised learning setup, as there is no training time series, methods that require training cannot be applied to unsupervised anomaly detection datasets. 
The problem statement for the \textit{supervised} variant is as follows:

\begin{prob} 
Given a training time series~$T_\text{train} \in \mathbb{R}^{n_\text{train} \times d}$, the associated ground truth labels~$\mathbf{y}_\text{train} \in \mathbb{B}^{n_\text{train}}$, and a test time series~$T_\text{test} \in \mathbb{R}^{n_\text{test} \times d}$, the objective of \textit{supervised anomaly detection} is to output an anomaly score vector~$\hat{\mathbf{y}}_\text{test} \in \mathbb{R}^{n_\text{test}}$ where $\hat{\mathbf{y}}_\text{test}[i]$ stores the score associated with the $i$-th time step. 
\end{prob}

The supervised learning setup is akin to the binary classification problem, with the key difference being that the output during test time is anomaly scores instead of binary labels. 
The problem statement for the \textit{semi-supervised} variant is as follows:

\begin{prob} 
Given a training time series~$T_\text{train} \in \mathbb{R}^{n_\text{train} \times d}$ consisting solely of normal data and a test time series~$T_\text{test} \in \mathbb{R}^{n_\text{test} \times d}$, the goal of \textit{semi-supervised anomaly detection} is to output an anomaly score vector~$\hat{\mathbf{y}}_\text{test} \in \mathbb{R}^{n_\text{test}}$ where $\hat{\mathbf{y}}_\text{test}[i]$ stores the score associated with the $i$-th time step. 
\end{prob}

The formulation of this semi-supervised learning setup is similar to the one-class classification problem, but the output during test time is an anomaly scores instead of a binary labels.

%% file: section/method.tex
\section{Methodology}
\label{sec:method}
This section details the application of the Matrix Profile (MP) to the multidimensional time series anomaly detection (TSAD) problem.
First, we discuss various strategies for creating a multidimensional MP from a pairwise distance tensor. 
Following this, we illustrate how to leverage the multidimensional MP in different anomaly detection problem formulations.

\subsection{Multidimensional MP}
\label{sec:multi_mp}
Before delving into the strategies for summarizing pairwise distance tensors from multidimensional time series, we will first demonstrate how MP summarizes the pairwise distance matrix from univariate time series. 
This approach will acquaint readers with both the MP concept and the graphical representation utilized to explain the summarization operations. 
Fig.~\ref{fig:mp_vanilla} illustrates the summarization process of the original MP for univariate time series~\cite{yeh2016matrix}.

\input{insert/fig_mp_vanilla}

\sloppy
In Fig.~\ref{fig:mp_vanilla}, we present a pairwise distance matrix~$D \in \mathbb{R}^{(n_1 - m + 1) \times (n_2 - m + 1)}$ generated from two time series~$\mathbf{t}_1 \in \mathbb{R}^{n_1}$ and $\mathbf{t}_2 \in \mathbb{R}^{n_2}$ with a subsequence length of~$m$. 
The MP of $\mathbf{t}_1$ joined with $\mathbf{t}_2$ is defined as the distance between each subsequence in $\mathbf{t}_1$ and its nearest neighbor in $\mathbf{t}_2$. 
To compute such a MP from $D$, we simply identify the minimal value from each column in $D$. 
The $i$-th column of $D$ consists of the distances between the $i$-th subsequence in $\mathbf{t}_1$ and each subsequence in $\mathbf{t}_2$. 
Therefore, the minimal value corresponds to the nearest neighbor distance.

As demonstrated in Fig.~\ref{fig:kn_problem}, the approach of simply summing the distances computed using each individual dimension will not work, as it would obscure the anomalous pattern amidst all the dimensions that consist of normal patterns. 
This problem is known as the $K$ of $N$ anomaly detection problem, introduced in Section~\ref{sec:intro}. 
To address this, let us consider the multidimensional scenario where we have a pairwise distance tensor~$\mathbf{D} \in \mathbb{R}^{(n_1 - m + 1) \times (n_2 - m + 1) \times d}$, generated from two time series~$T_1 \in \mathbb{R}^{n_1 \times d}$ and $T_2 \in \mathbb{R}^{n_2 \times d}$ with a subsequence length of~$m$. 
Generally, the most anomalous dimension can be identified by sorting the dimension-wise distances between a subsequence and its nearest neighbor. 
The dimension with the largest distance value is deemed the most anomalous (as it has the largest nearest neighbor distance), and vice versa~\cite{yeh2017matrix,tafazoli2023matrix}. 
Fig.~\ref{fig:dim_sort} illustrates this: given two four-dimensional subsequences $T_i$ and $T_j$, sorting the dimensions based on the dimension-wise distances reveals that the fourth dimension is the most anomalous.

\input{insert/fig_dim_sort}

Given that we need to incorporate a sorting mechanism into MP computation process, we have two options: 1) a \textit{post-sorting} design, where the sorting algorithm is applied after the nearest neighbors are identified, as demonstrated in Fig.~\ref{fig:post_sort}, and 2) a \textit{pre-sorting} design, where the sorting algorithm is applied before the nearest neighbors are identified, as demonstrated in Fig.~\ref{fig:pre_sort}.

\input{insert/fig_mp_post_sort}

The post-sorting strategy was first outlined in~\cite{tafazoli2023matrix}. 
It utilizes a sorting algorithm after the nearest neighbor is identified for each dimension independently. 
Following the example shown in Fig.~\ref{fig:post_sort}, after the nearest neighbor distances are identified (i.e., the ``find nearest neighbor" step in Fig.~\ref{fig:post_sort}), we obtain the MP in a size $(n_1 - m + 1) \times d$ matrix, where each size $n_1 - m + 1$ vector is the univariate MP for one of the dimensions. 
Next, the sorting algorithm is applied to the $(n_1 - m + 1) \times d$ matrix at each time step. In other words, the $O(d \log{d})$ sorting operation is applied $n_1 - m + 1$ times. 
Therefore, the overall time complexity for applying the post-sorting strategy is $O(n_1 d \log{d})$.

\input{insert/fig_mp_pre_sort}

The pre-sorting strategy, as illustrated in Fig.~\ref{fig:pre_sort}, was first adopted in~\cite{yeh2017matrix} for time series motif discovery. 
To apply the pre-sorting strategy to the anomaly detection problem, we apply the $O(d \log{d})$ sorting operation to the distances associated with each pair of subsequences. 
In other words, the sorting operation is applied to the pairwise distance tensor $(n_1 - m + 1) \times (n_2 - m + 1)$ times. 
After the sorting is completed, we use the sorted pairwise distance matrix to find the nearest neighbor for each subsequence in $\mathbf{t}_1$. 
The output MP is a matrix of size $(n_1 - m + 1) \times d$. 
The overall time complexity for applying the pre-sorting strategy is $O(n_1 n_2 d \log{d})$.

We made two modifications to the procedure outlined in~\cite{yeh2017matrix} to adapt it for anomaly detection. 
The first modification involves reversing the order of the sorting algorithm. 
In the original application in~\cite{yeh2017matrix}, for motif discovery, the sorting goal is to find the dimension with the smallest nearest neighbor distance. 
In contrast, our goal for anomaly detection is to find the dimension with the largest nearest neighbor distance. 
Therefore, we sort the dimensions from largest to smallest distance value. 
The second modification is the removal of the \textit{cumulative sum} step, which is unnecessary for anomaly detection, as shown in~\cite{tafazoli2023matrix}.

\sloppy
For both strategies, if the input time series are $d$-dimensional, the associated multidimensional MP will also be $d$-dimensional. 
The $i$-th dimension in the MP can be used to detect anomalies that span \textit{at least} $i$ dimensions. 
In Fig.~\ref{fig:mmp_dim}, we illustrate a four-dimensional time series with four inserted anomalies: a one-dimensional (1D) anomaly, a 2D anomaly, a 3D anomaly, and a 4D anomaly.

\input{insert/fig_mmp_dim}

In the figure, for both strategies, the first dimension of the multidimensional MP can be used to detect all four anomalies, the second dimension can be used to detect the last three anomalies, the third dimension can be used to detect the last two anomalies, and the fourth dimension can only be used to detect the four-dimensional anomaly. 
The MP shown in Fig.~\ref{fig:kn_problem} is the first dimension of the multidimensional MP. 
Note that if the user wants to find all anomalies that span \textit{at least} one dimension, the $O(d \log{d})$ sorting operation can be replaced with an $O(d)$ maximum value finding operation. 
This modification reduces the time complexity of the post-sorting strategy to $O(n_1 d)$ and the time complexity of the pre-sorting strategy to $O(n_1 n_2 d)$.

There are two major differences between the two strategies: 1) time complexity and 2) anomaly detection capability. 
To illustrate the differences in terms of time complexity, we present Table~\ref{tab:mmp_time}, which shows the time complexities associated with computing the multidimensional MP. 
We assume that the pairwise distance tensor is computed exactly using the algorithm presented in~\cite{zhu2016matrix}. 
Note that ``post-max" and ``pre-max" are variants of ``post-sorting" and ``pre-sorting", respectively, where the sorting operation is replaced with a max operation (or selection operation like the introselect~\cite{musser1997introspective} algorithm).

\input{insert/tab_mmp_time}

The pre-sorting-based variants (i.e., pre-sorting and pre-max) are more computationally expensive than their post-sorting-based counterparts (i.e., post-sorting and post-max). 
This is because all the pre-sorting-based variants operate on the pairwise distance tensor, while the post-sorting-based variants operate on the summary of the pairwise distance tensor. 
Considering the time complexity of computing the pairwise distance tensor, we can see that the pre-max, post-sorting, and post-max strategies do \textit{not} increase the overall time complexity in terms of big O notation. 
However, the pre-sorting strategy increases the overall time complexity by a factor of $\log{d}$.
As we utilize the algorithm presented in~\cite{zhu2016matrix}, the space complexity for all variants is $O(\max(n_1, n_2)d)$. 

We compare the actual runtime of the four variants listed in Table~\ref{tab:mmp_time}. 
Since the runtime of these algorithms depends solely on the scale of the data, we generate random time series and measure the runtime of each variant. 
The time complexity depends on $d$, $n_1$, and $n_2$, so we measure the average runtime across 16 trials under different settings of these variables. 
When altering $d$, we maintain $n_1$ and $n_2$ at $2^{12}$. 
Conversely, when adjusting $n_1$ and $n_2$, we keep $d$ constant at 64. 
The runtime is presented in Fig.~\ref{fig:prepost_time}, and the results align with the time complexity presented in Table~\ref{tab:mmp_time}.

\input{insert/fig_prepost_time}

In terms of anomaly detection capability, there are also differences between the two strategies. 
Specifically, the MP generated with the post-sorting strategy cannot detect anomalies caused by changes in correlation between dimensions. 
In Fig~\ref{fig:mmp_diff}, we present a three-dimensional time series with an inserted anomalous subsequence within one of the dimensions. 
The inserted subsequence is considered an anomaly because, outside of the inserted anomaly, the correlation between the modified dimension and the other two dimensions is close to one.

\input{insert/fig_mmp_diff}

As demonstrated in the figure, the MP generated using the post-sorting strategy cannot detect this type of anomaly. 
The post-sorting strategy fails to detect such anomalies because the correlation information is lost when identifying the nearest neighbors. 
Since the pre-sorting strategy sorts the pairwise distance tensor, the correlation information is utilized during the sorting operation.

In conclusion, the pre-sorting-based variants are generally more desirable for anomaly detection because they can detect a wider range of anomalies. 
However, the post-sorting-based variants could be a better option in two scenarios: 1) when the efficiency of the anomaly detection system is critical for the application, or 2) when there are no anomalies defined by changes in cross-dimension correlation in the data.

\subsection{MP for Anomaly Detection}
To apply MP to multidimensional TSAD datasets, we initially obtain the MP from the time series in the dataset. 
This is achieved by performing either a self-join or AB-join, depending on the learning setups. 
Given that the anomaly score is a one-dimensional time series, it is necessary to reduce the multidimensional MP to a one-dimensional time series. 
This is accomplished by selecting one dimension from the multidimensional MP as the anomaly score curve for subsequent processing. 
We then post-process the anomaly score curve with a moving average to enhance temporal consistency. 
In this section, we also explore two additional aspects of applying the multidimensional MP to anomaly detection problems:
1) The extension of MP to identify the $k$-th nearest neighbor, which can boost the performance of MP for anomaly detection. 
2) The determination of the type of join to be performed for various anomaly detection learning setups.

\subsubsection{$k$-Nearest Neighbor Extension}
To further improve the performance of MP for anomaly detection, we modify MP by extending the ``find nearest neighbor" operation to be capable of finding the $k$-th nearest neighbor. 
This capability is crucial as some datasets, despite the rarity of anomalies, exhibit repeatable patterns within these anomalies~\cite{lu2022matrix}.
The original ``find nearest neighbor" operation has a time complexity of $O(n_1 n_2 d)$ in both Fig.~\ref{fig:post_sort} and Fig.~\ref{fig:pre_sort} because it needs to traverse the tensor to find the nearest neighbor. 
In theory, extending the operation to find the $k$-th nearest neighbor could be implemented with the introselect~\cite{musser1997introspective} or QuickSelect algorithm~\cite{hoare1961algorithm}, which would not increase the time complexity compared to the one-nearest neighbor variant. 
However, since we need to avoid trivial matches~\cite{yeh2016matrix} of closer neighbors (i.e., the trivial matches of the first to $(k-1)$-th neighbors) when finding the $k$-th nearest neighbor, we cannot simply apply these efficient algorithms.

For this reason, we propose an efficient $k$-th nearest neighbor selection algorithm that takes into account trivial matches. 
Before discussing the proposed algorithm, let us introduce two alternative naive solutions with different time complexities: 1) a brute force solution (Algorithm~\ref{alg:find_knn_0}), where the linear time minimum value search algorithm is applied $k$ times, and 2) a naive sorting-based solution (Algorithm~\ref{alg:find_knn_1}), where the array is first sorted before acquiring the $k$-th nearest neighbor. 
The first solution increases the time complexity of the ``find nearest neighbor" operation from $O(n_1 n_2 d)$ to $O(k n_1 n_2 d)$, while the second solution increases the time complexity to $O(n_1 n_2 \log{n_2} d)$. 
If a large $k$ is used, the second solution is faster. 
Conversely, if a small $k$ is used, the first solution is faster.

\input{insert/alg_find_knn_0}

Algorithm~\ref{alg:find_knn_0} presents the brute force solution. 
The algorithm is largely self-explanatory, with the exception of \texttt{line 8} where the exclusion zone is applied to the $\mathbf{d}_{ij}$ array. 
The purpose of this exclusion zone is to prevent the subsequence starting at index $i_\text{min}$ and its trivial matches from being rediscovered in future iterations. 
This is implemented with $\mathbf{d}_{ij}[i_\text{min} - \frac{m}{2}:i_\text{min} + \frac{m}{2}] \gets \inf$.

\input{insert/alg_find_knn_1}

Algorithm~\ref{alg:find_knn_1} presents the naive sorting-based solution. 
A sorting algorithm is used in \texttt{line 5} to determine the order of subsequences based on their distance values. 
In \texttt{line 8}, an if statement is used to check whether the current subsequence is a trivial match of one of the previously identified nearest neighbors. 
In \texttt{line 10}, the algorithm checks if the $k$-th nearest neighbor has been identified.

The proposed efficient $k$-th nearest neighbor selection algorithm extends Algorithm~\ref{alg:find_knn_1} by replacing \texttt{line 5} with \texttt{line 5}$_\text{new}$, \texttt{line 6}$_\text{new}$, and \texttt{line 7}$_\text{new}$ in Algorithm~\ref{alg:find_knn_2}. 
In this extension, the input array to the $\textsc{ArgSort}(\cdot)$ function in \texttt{line 5} of Algorithm~\ref{alg:find_knn_1} is reduced from a $n_2$ array to a $k m$ array using a linear time selection algorithm. 
This algorithm selects the most similar $k m$ subsequences, where $m$ represents both the subsequence length and the length of the exclusion zone for trivial matches. 
Given that the $k$-th nearest neighbor is guaranteed to be within the set of the most similar $k m$ subsequences, we only need to sort $k m$ items in \texttt{line 6}$_\text{new}$ and \texttt{line 7}$_\text{new}$. 
Assuming $n_2 \ggg k m \log{k m}$, the time complexity of the proposed method is $O(n_1 n_2 d)$.

\input{insert/alg_find_knn_2}

The time complexities of the three $k$NN finding algorithms are $O(k n_1 n_2 d)$, $O(n_1 n_2 \log{n_2} d)$, and $O(n_1 n_2 d)$ for brute force, naive sorting, and the proposed method, respectively. 
The key variables that differentiate these three algorithms in terms of time complexity are $k$ and $n_2$. 
We generate random time series for runtime experiments. 
The default settings for the experiments are $k=64$, $n_1=16$, $n_2=2^{14}$, and $d=4$. 
When testing different $k$s, we fix the other settings to the default value. 
Similarly, when testing different $n_2$s, we keep the other settings at their default values. 
The reported runtime is an average computed over 100 trials.
The experiment results are shown in Fig.~\ref{fig:findnn_time}. 
The proposed method outperforms the alternative methods.

\input{insert/fig_findnn_time}

\subsubsection{Type of Join}
Different types of joins need to be used to compute the MP for different learning settings (i.e., unsupervised, supervised, and semi-supervised) in anomaly detection.

For unsupervised learning datasets, each dataset contains only one test time series. 
Therefore, the MP for this learning setup is computed using a self-join on the test time series.

For supervised learning datasets, each dataset contains a test time series, a training time series, and the ground truth label associated with the training time series. 
We concatenate the training time series with the test time series, then compute MP using a self-join on the concatenated time series. 
In other words, we compute MP associated with the training time series using both the training and test time series. 
Similarly, we compute MP associated with the test time series using both the training and test time series. 
There are two reasons for this design. 
First, the purpose of the training time series and its label is for hyper-parameter tuning. 
Including the test time series in the computation of MP for the training time series can improve the accuracy of test performance estimation. 
Second, including the training time series when computing MP for the test time series can enhance performance due to the utilization of patterns in both time series.

For semi-supervised learning datasets, each dataset comprises a test time series and a training time series. 
The training time series consists solely of normal patterns. 
Therefore, patterns that exist only in the test time series, and not in the training time series, are likely anomalous. 
We can apply an AB-join to join the test time series with the training time series to identify such patterns. 
Consequently, we utilize AB-join for anomaly detection in semi-supervised learning datasets.

%% file: insert/fig_mp_vanilla.tex
\begin{figure}[htp]
\centerline{
\includegraphics[width=0.65\linewidth]{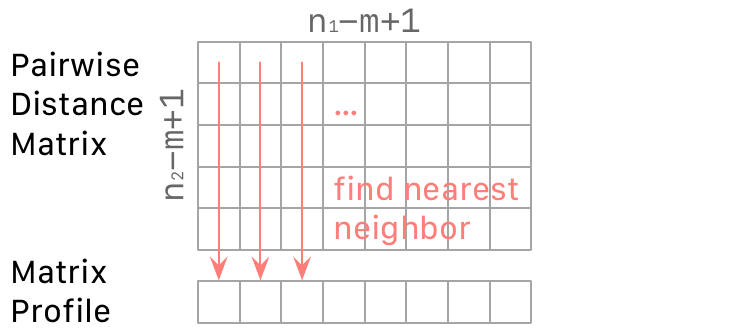}
}
\caption{
The Matrix Profile summarizes the pairwise distance matrix of a pair of univariate time series by identifying the nearest neighbor within the matrix.
}
\label{fig:mp_vanilla}
\end{figure}

%% file: insert/fig_dim_sort.tex
\begin{figure}[htp]
\centerline{
\includegraphics[width=0.9\linewidth]{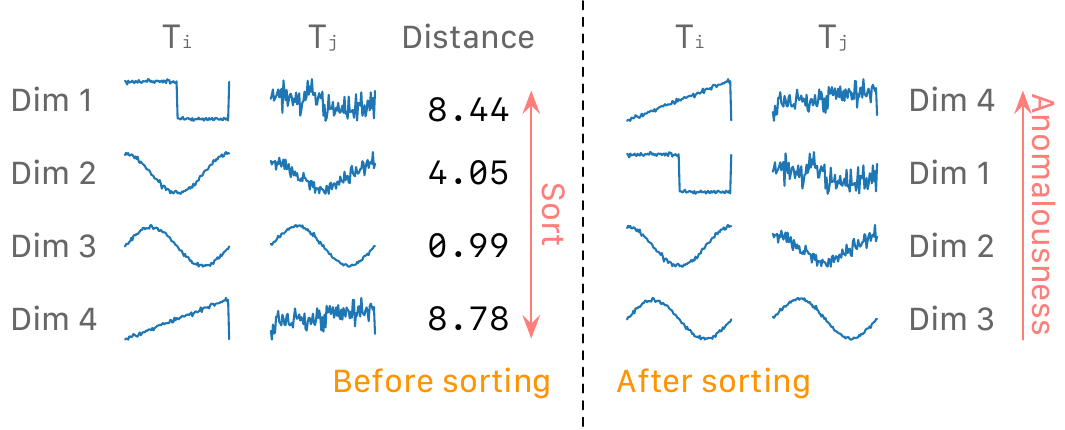}
}
\caption{
In this example, based on the sorted results, the anomaly is most likely contained within the first and fourth dimensions.
}
\label{fig:dim_sort}
\end{figure}

%% file: insert/fig_mp_post_sort.tex
\begin{figure}[htp]
\centerline{
\includegraphics[width=0.65\linewidth]{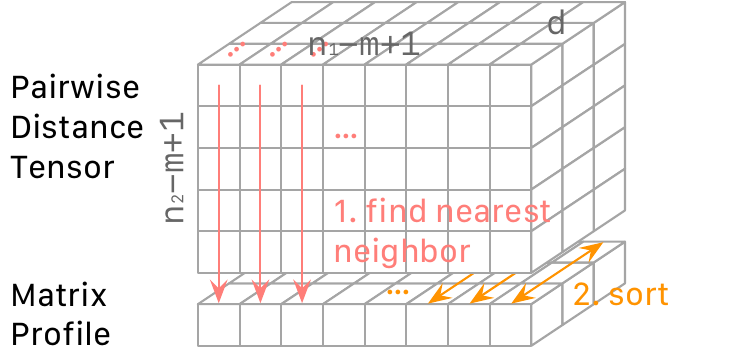}
}
\caption{
The Matrix Profile uses the post-sorting strategy to summarize the pairwise distance tensor from a pair of multidimensional time series.
}
\label{fig:post_sort}
\end{figure}

%% file: insert/fig_mp_pre_sort.tex
\begin{figure}[htp]
\centerline{
\includegraphics[width=0.65\linewidth]{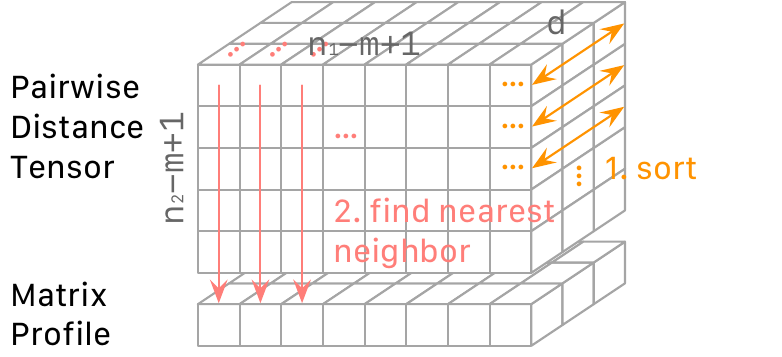}
}
\caption{
The Matrix Profile uses the pre-sorting strategy to summarize the pairwise distance tensor from a pair of multidimensional time series.
}
\label{fig:pre_sort}
\end{figure}

%% file: insert/fig_mmp_dim.tex
\begin{figure}[htp]
\centerline{
\includegraphics[width=0.99\linewidth]{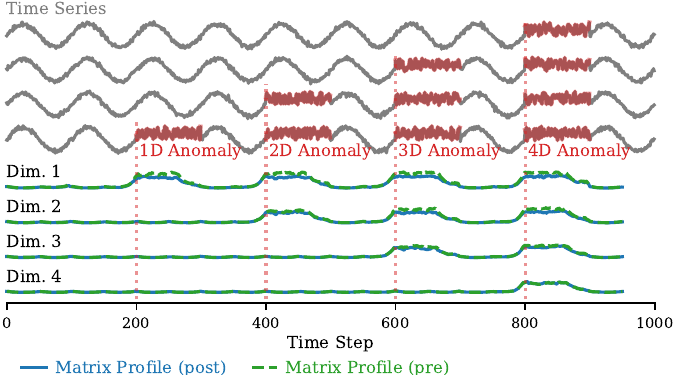}
}
\caption{
Different dimensions of the multidimensional Matrix Profile can be used to detect anomalies that span different numbers of dimensions.
}
\label{fig:mmp_dim}
\end{figure}

%% file: insert/tab_mmp_time.tex
\begin{table}[htp]
\caption{The time complexities associated with computing various variants of the multidimensional Matrix Profile.}
\label{tab:mmp_time}
\begin{center}
\footnotesize
\begin{tabular}{l||cc|c}
\shline
Variant & Distance tensor & Sort or max & Overall \\ \hline \hline
Post-sorting & $O(n_1 n_2 d)$ & $O(n_1 d \log{d})$ & $O(n_1 n_2 d)$ \\ 
Post-max & $O(n_1 n_2 d)$ & $O(n_1 d)$ & $O(n_1 n_2 d)$ \\ 
Pre-sorting & $O(n_1 n_2 d)$ & $O(n_1 n_2 d \log{d})$ & $O(n_1 n_2 d \log{d})$ \\ 
Pre-max & $O(n_1 n_2 d)$ & $O(n_1 n_2 d)$ & $O(n_1 n_2 d)$ \\ \shline
\end{tabular}
\end{center}
\end{table} 

%% file: insert/fig_prepost_time.tex
\begin{figure}[htp]
\centerline{
\includegraphics[width=0.99\linewidth]{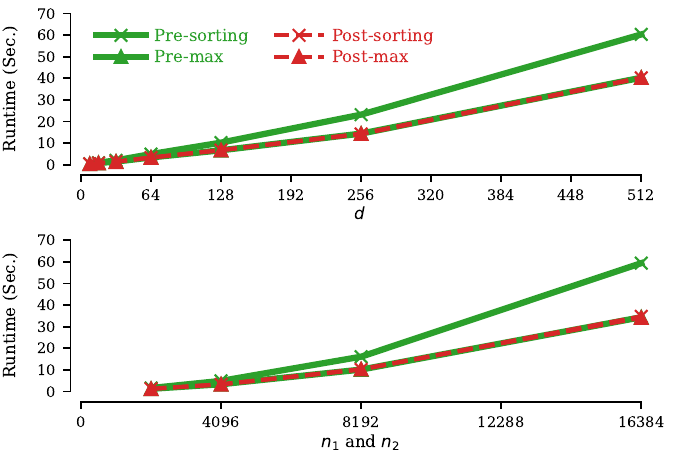}
}
\caption{
This figure illustrates how the four variants of the multidimensional Matrix Profile scale with the number of dimensions ($d$) and the length of the time series ($n_1$ and $n_2$). 
The runtimes for pre-max, post-sorting, and post-max are nearly identical, while pre-sorting is slower in comparison.
}
\label{fig:prepost_time}
\end{figure}

%% file: insert/fig_mmp_diff.tex
\begin{figure}[htp]
\centerline{
\includegraphics[width=0.99\linewidth]{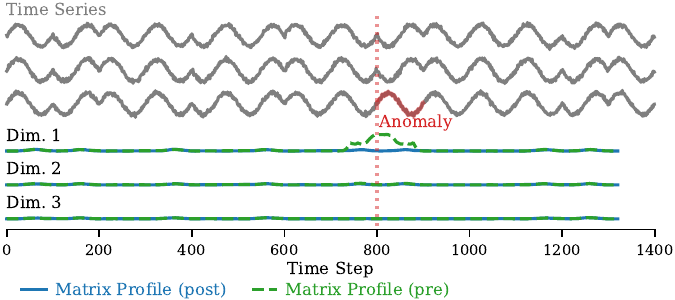}
}
\caption{
This figure presents the pre-sorting and post-sorting multidimensional Matrix Profile variants for a time series with an inserted anomaly. 
Only the pre-sorting variant detects the inserted anomaly.
}
\label{fig:mmp_diff}
\end{figure}

%% file: insert/alg_find_knn_0.tex
\begin{algorithm}[htp]
    \centering
    \caption{Brute Force $k$NN Finding Algorithm\label{alg:find_knn_0}}
    \footnotesize
    \begin{algorithmic}[1]
        \Input{pairwise distance tensor~$\mathbf{D} \in \mathbb{R}^{(n_1 - m + 1) \times (n_2 - m + 1) \times d}$, number of neighbors~$k \in \mathbb{N}$}
        \Output{$k$-th nearest neighbor index~$I \in \mathbb{N}^{(n_1 - m + 1) \times d}$}
        \Function{Find$k$NN}{$\mathbf{D}, k$}
        \State $I \gets $ zero matrix with size $(n_1 - m + 1) \times d$
        \For{$i \in [0, \cdots, n_1 - m + 1]$}
        \For{$j \in [0, \cdots, d]$}
        \State $\mathbf{d}_{ij} \gets \mathbf{D}[:, i, j]$
        \For{$l \in [0, \cdots, k]$}
        \State $i_\text{neighbor} \gets$ \textsc{FindMinIndex}$(\mathbf{d}_{ij})$ \Comment{$O(n_2)$}
        \State $\mathbf{d}_{ij} \gets$ \textsc{ApplyExclusionZone}$(\mathbf{d}_{ij}, i_\text{min})$
        \EndFor
        \State $I[i, j] \gets i_\text{neighbor}$
        \EndFor
        \EndFor
        \State \Return I
        \EndFunction
    \end{algorithmic}
\end{algorithm}

%% file: insert/alg_find_knn_1.tex
\begin{algorithm}[htp]
    \centering
    \caption{Naive Sorting-based $k$NN Finding Algorithm\label{alg:find_knn_1}}
    \footnotesize
    \begin{algorithmic}[1]
        \Input{pairwise distance tensor~$\mathbf{D} \in \mathbb{R}^{(n_1 - m + 1) \times (n_2 - m + 1) \times d}$, number of neighbors~$k \in \mathbb{N}$}
        \Output{$k$-th nearest neighbor index~$I \in \mathbb{N}^{(n_1 - m + 1) \times d}$}
        \Function{Find$k$NN}{$\mathbf{D}, k$}
        \State $I \gets $ zero matrix with size $(n_1 - m + 1) \times d$
        \For{$i \in [0, \cdots, n_1 - m + 1]$}
        \For{$j \in [0, \cdots, d]$}
        \State $\mathbf{i}_\text{neighbors} \gets$ \textsc{ArgSort}$(\mathbf{D}[:, i, j])$ \Comment{$O(n_2 \log{n_2})$}
        \State $l \gets 0$
        \For{$i_\text{neighbor} \in \mathbf{i}_\text{neighbors}$}
        \If{$i_\text{neighbor}$ is a trivial match} continue
        \EndIf 
        \State $l \gets l + 1$
        \If{$l = k$} break
        \EndIf
        \EndFor
        \State $I[i, j] \gets i_\text{neighbor}$
        \EndFor
        \EndFor
        \State \Return I
        \EndFunction
    \end{algorithmic}
\end{algorithm}

%% file: insert/alg_find_knn_2.tex

\algrenewcommand\alglinenumber[1]{#1$_\text{new}$}

\begin{algorithm}[htp]
    \centering
    \caption{Proposed Efficient $k$NN Finding Algorithm\label{alg:find_knn_2}}
    \footnotesize
    \begin{algorithmic}[1]
        \setcounter{ALG@line}{4}
        \State $\mathbf{i}_\text{neighbors} \gets$ \textsc{ArgSelect}$(\mathbf{D}[:, i, j], k \times m)$ \Comment{$O(n_2)$}
        \State $\mathbf{i}_\text{order} \gets$ \textsc{ArgSort}$(\mathbf{D}[\mathbf{i}_\text{neighbors}, i, j])$ \Comment{$O(k m \log{k m})$}
        \State $\mathbf{i}_\text{neighbors} \gets \mathbf{i}_\text{neighbors}[\mathbf{i}_\text{order}]$
    \end{algorithmic}
\end{algorithm}

\algrenewcommand\alglinenumber[1]{#1}

%% file: insert/fig_findnn_time.tex
\begin{figure}[htp]
\centerline{
\includegraphics[width=0.99\linewidth]{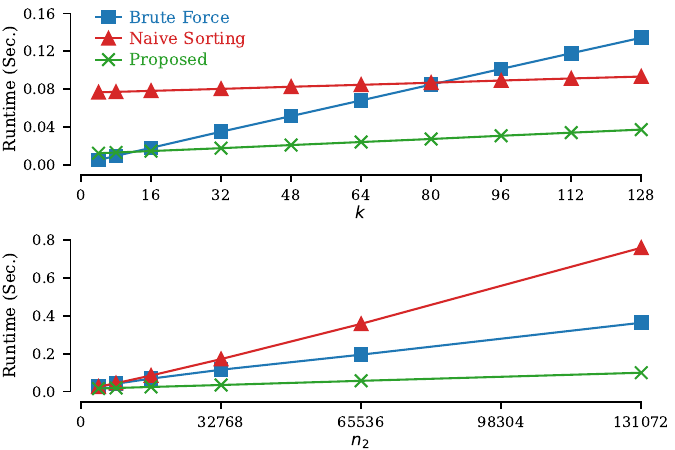}
}
\caption{
This figure demonstrates how different $k$NN finding algorithms scale with the parameters $k$ and time series length~$n_2$.
The proposed $k$NN finding algorithm is more efficient compared to the baseline methods.
}
\label{fig:findnn_time}
\end{figure}

%% file: section/experiment.tex
\section{Experiment}
\label{sec:experiment}
In this section, we first outline the datasets used for the experiments in Section~\ref{sec:dataset}.
We then discuss the experimental protocol for different learning setups, such as unsupervised, supervised, and semi-supervised, in Section~\ref{sec:exp_setting}.
The baseline methods for comparison are detailed in Section~\ref{sec:baseline}, and the benchmark results, comparing the MP-based method with the baseline methods, are presented in Section~\ref{sec:result}.
As there are no labeled training data for both unsupervised and semi-supervised learning datasets, the hyper-parameters are set manually.
A study on these hyper-parameters is presented in Section~\ref{sec:hparameter}.
While this section details our evaluation across 119 datasets, we have also evaluated our approach against the TSB-AD benchmark~\cite{yeh2026matrix}. 
The full MMPAD source code and these expanded TSB-AD results are available in our open-source repository~\cite{yeh2026mmpad}, while further details regarding the paper's experiments can be found in~\cite{supplementary}.

\subsection{Dataset}
\label{sec:dataset}
Experiments were conducted on multidimensional time series anomaly detection (TSAD) datasets, which are available on the supporting website of~\cite{wenig2022timeeval}. 
Our experimental setup made use of 48 unsupervised learning datasets, 25 supervised datasets, and 46 semi-supervised datasets. 
These 119 datasets, which are derived from eight different dataset collections, are detailed in Table~\ref{tab:data}.
The dimensionality of these datasets varies from 2 to 38.
The total number of time steps across all 119 datasets is 8,849,589, and the total number of values is 71,998,980.

\input{insert/tab_dataset}

\subsection{Experiment Setting}
\label{sec:exp_setting}
\sloppy
The experiment settings are determined based on the problem formulations associated with each of the learning setups (i.e., unsupervised, supervised, and semi-supervised) introduced in Section~\ref{sec:problem}. 
The process for measuring performance is consistent across all three learning setups. 
The specific performance functions utilized in this paper are AUC-ROC~\cite{hanley1982meaning,bradley1997use} and AUC-P$_{\text{T}}$R$_{\text{T}}$~\cite{tatbul2018precision}. 
These are commonly used threshold-agnostic performance measures for TSAD problems~\cite{schmidl2022anomaly}.

For supervised learning datasets, the hyper-parameters for the MP-based method are determined based on the results of a hyper-parameter search using the training data. 
However, due to the absence of training labels for both unsupervised and semi-supervised datasets, the hyper-parameters are manually set. 
For unsupervised datasets, we set the subsequence length to 64, adopt the pre-max variant, use the first dimension of the multidimensional MP, and set $k$ for $k$-th nearest neighbor finding to 15. 
For semi-supervised datasets, we use a similar set of hyper-parameters as the unsupervised datasets, but set the $k$ to one instead of 15. 
The process of setting the hyper-parameters will be further discussed in Section~\ref{sec:hparameter}.

\subsection{Baseline Method}
\label{sec:baseline}
We included 19 baseline methods for comparison with the MP-based method. 
We report the performances of baseline methods that are available on at least 50\% of the datasets within a learning setup. 
The methods included in our benchmark, along with their availability in different learning settings, are shown in Table~\ref{tab:baseline}. 
We use check marks to highlight the methods that meet the availability criteria.

\input{insert/tab_baseline_1}

\subsection{Benchmark Result}
\label{sec:result}
The benchmark results for the MP-based anomaly detection system, along with the 19 baseline methods, are shown in Table~\ref{tab:benchmark}. 
Following~\cite{schmidl2022anomaly}, we use box plots to summarize the AUC-ROC and AUC-P$_{\text{T}}$R$_{\text{T}}$ across different datasets. 
For each method, we create a separate box plot for each learning setup.

\input{insert/tab_benchmark_1}

Of the 12 methods that work for unsupervised learning datasets, $k$-Means is the most competitive in terms of both AUC-ROC and AUC-P$_{\text{T}}$R$_{\text{T}}$. 
When comparing MP with $k$-Means~\cite{yairi2001fault}, it is evident that $k$-Means outperforms MP. 
However, the difference in performance distribution is \textit{not} significant, according to paired samples t-tests with $\alpha=0.05$. 
MP outperforms the other 11 methods in both AUC-ROC and AUC-P$_{\text{T}}$R$_{\text{T}}$. 
Thus, both MP and $k$-Means are competitive for unsupervised learning datasets.

For supervised learning datasets, only two baseline methods are available: HIF~\cite{marteau2017hybrid} and MultiHMM~\cite{li2017multivariate}. 
MP outperforms both of these baseline methods, making it the most competitive method for supervised learning datasets.

For semi-supervised learning datasets, the most competitive baseline method is Telemanom~\cite{hundman2018detecting}. 
Although Telemanom outperforms MP, the latter exhibits similar AUC-P$_{\text{T}}$R$_{\text{T}}$ to Telemanom according to paired samples t-tests with $\alpha=0.05$. 
While the MP may not be the best method for semi-supervised learning datasets, its performance remains competitive when compared to the best method.

Out of the 20 methods presented in Table~\ref{tab:benchmark}, only nine are applicable to datasets of more than one learning setup. 
These nine methods are the MP, EIF~\cite{hariri2019extended}, iForest~\cite{liu2008isolation}, $k$NN~\cite{ramaswamy2000efficient}, CBLOF~\cite{he2003discovering}, LOF~\cite{breunig2000lof}, COPOD~\cite{li2020copod}, PCC~\cite{shyu2003novel}, and HBOS~\cite{goldstein2012histogram}. 
Among these nine methods, MP is the only one that is both competitive and available to all three learning setups.

The runtime for the MP-based anomaly detection system is depicted in Fig.~\ref{fig:ad_runtime}.
Given that MP is an embarrassingly parallelizable method, we display the total runtime required to complete the benchmark datasets under various settings for the number of parallel jobs (or processes).
When the number of processes is set to 16, approximately two hours and 22 minutes are needed to compute all experiments.
Our method is implemented in Python 3.10.11 with numpy 1.24.3.
The experiment is performed on a Linux server equipped with AMD EPYC 7513 32-Core Processors.

\input{insert/fig_ad_runtime}

\subsection{Hyper-Parameter Study}
\label{sec:hparameter}
There are four major hyper-parameters associated with a MP-based multidimensional TSAD system: 1) subsequence length, 2) the value of $k$ for $k$-th nearest neighbor finding, 3) pre or post-sorting, and 4) the dimension selection strategy from the multidimensional MP. 
For the supervised learning setup, we can experiment with different hyper-parameter settings to select the one that yields the best training data performance. 
However, in the case of unsupervised and semi-supervised learning setups, where labeled training data is not available, these hyper-parameters need to be manually set.

In this section, we will discuss how to set these hyper-parameters and present experimental results associated with varying them. 
Additionally, we will demonstrate the effect of post-processing on anomaly detection performance. 
For the experiments, we set the hyper-parameters to their default settings and vary one at a time, observing the resultant change in performance.

Although the subsequence length is a non-sensitive hyper-parameter for MP~\cite{yeh2016matrix,lu2023matrix}, it remains crucial. 
This parameter must be set based on the data, such as the periodicity of the time series, which makes it impossible to provide a default value. 
The best strategy for setting this hyper-parameter would rely on either domain knowledge or exploratory data analysis~\cite{lu2023matrix}. 
In Fig~\ref{fig:param_sub}, we demonstrate the effect of setting the subsequence length to different values on the average performances.

\input{insert/fig_param_sub}

We observe that performance varies greatly when the subsequence length is small, but the rate of change in performance slows down and eventually plateaus as the subsequence length increases. 
By setting the subsequence length to 64, the MP-based system can achieve great performance for the specific benchmark datasets we used in our experiments. 
Overall, this observation aligns with the conclusion from previous papers that this hyper-parameter is non-sensitive, yet crucial.

Similar to the subsequence length, the hyper-parameter $k$ for finding the $k$-th nearest neighbor also depends on the dataset. 
In Fig~\ref{fig:param_knn}, we illustrate the impact of setting $k$ to different values on the average performances.

\input{insert/fig_param_knn}

For semi-supervised learning datasets, setting $k$ to one typically yields better performance, as the training data does not contain anomalies (i.e., the twin freak problem does not occur). 
For unsupervised learning, setting $k$ to a value greater than one tends to improve performance, as the test time series may contain a small number of anomalies that recur. 
For the unsupervised learning datasets included in our benchmark, setting $k$ to 15 generally results in better performance.

In terms of choosing between pre-sorting or post-sorting variants, we recommend following the guidance provided in Section~\ref{sec:multi_mp} and opting for the pre-sorting variants, as they can identify a broader range of anomalies. 
Our experimental results, shown in Table~\ref{tab:param}, also support this decision, as the pre-sorting variant outperforms the post-sorting variant in terms of average performance.

\input{insert/tab_param_1}

When selecting a dimension from the multidimensional MP, the first dimension should be used as it can identify all sub-dimensional anomalies, as demonstrated in Fig.~\ref{fig:mmp_dim}. 
We compared this strategy to an alternative approach where the dimensions are reduced using the mean function, as implemented in~\cite{wenig2022timeeval}. 
Based on the experimental results shown in Table~\ref{tab:param}, using the first dimension outperforms the alternative. 
Lastly, as we demonstrate in Table~\ref{tab:param}, the moving average post-processing step generally enhances performance.

%% file: insert/tab_dataset.tex
\begin{table}[htp]
\caption{The number of datasets for each learning setup within the dataset collections.}
\label{tab:data}
\begin{center}
\resizebox{\linewidth}{!}{%
\begin{tabular}{l||ccc|c}
\shline
Collection & Unsupervised & Supervised & Semi-supervised & Total \\ \hline \hline
GutenTAG~\cite{wenig2022timeeval} & 23           & 23         & 23              & 69    \\
CalIt2~\cite{hutchins2006calit2}   & 1            & 0          & 0               & 1     \\
Daphnet~\cite{bachlin2009wearable}  & 3            & 0          & 0               & 3     \\
Exathlon~\cite{jacob2020exathlon} & 0            & 2          & 0               & 2     \\
Genesis~\cite{von2018anomaly}  & 1            & 0          & 0               & 1     \\
MITDB~\cite{goldberger2000physiobank}    & 4            & 0          & 0               & 4     \\
SMD~\cite{su2019robust}      & 0            & 0          & 23              & 23    \\
SVDB~\cite{goldberger2000physiobank}     & 16           & 0          & 0               & 16    \\ \hline
Total    & 48           & 25         & 46              & 119   \\ \shline
\end{tabular}
}
\end{center}
\end{table}

%% file: insert/tab_baseline_1.tex
\begin{table}[htp]
\caption{
The completion rates of various methods for different learning setups. 
Only methods with a completion rate of at least 50\% are included in the benchmark
}
\label{tab:baseline}
\begin{center}
\resizebox{\linewidth}{!}{%
\begin{tabular}{l|ccc}
\shline
\multirow{2}{*}{Method} & \multicolumn{3}{c}{Percent Complete}           \\
& Unsupervised & Supervised & Semi-supervised \\ \hline \hline
Matrix Profile & 100\% (\checkmark) & 100\% (\checkmark) & 100\% (\checkmark) \\
$k$-Means~\cite{yairi2001fault} & 85\% (\checkmark) & 0\% & 7\% \\
EIF~\cite{hariri2019extended} & 100\% (\checkmark) & 8\% & 50\% (\checkmark) \\
iForest~\cite{liu2008isolation} & 100\% (\checkmark) & 8\% & 50\% (\checkmark) \\
$k$NN~\cite{ramaswamy2000efficient} & 100\% (\checkmark) & 8\% & 50\% (\checkmark) \\
CBLOF~\cite{he2003discovering} & 100\% (\checkmark) & 8\% & 50\% (\checkmark) \\
IF-LOF~\cite{cheng2019outlier} & 100\% (\checkmark) & 8\% & 0\% \\
LOF~\cite{breunig2000lof} & 100\% (\checkmark) & 8\% & 50\% (\checkmark) \\
COPOD~\cite{li2020copod} & 100\% (\checkmark) & 8\% & 50\% (\checkmark) \\
PCC~\cite{shyu2003novel} & 100\% (\checkmark) & 8\% & 50\% (\checkmark) \\
HBOS~\cite{goldstein2012histogram} & 100\% (\checkmark) & 8\% & 50\% (\checkmark) \\
Torsk~\cite{heim2019adaptive} & 62\% (\checkmark) & 0\% & 28\% \\
COF~\cite{tang2002enhancing} & 56\% (\checkmark) & 0\% & 0\% \\
HIF~\cite{marteau2017hybrid} & 0\% & 100\% (\checkmark) & 0\% \\
MultiHMM~\cite{li2017multivariate} & 0\% & 52\% (\checkmark) & 0\% \\
Telemanom~\cite{hundman2018detecting} & 0\% & 0\% & 100\% (\checkmark) \\
OmniAnomaly~\cite{su2019robust} & 0\% & 0\% & 100\% (\checkmark) \\
DeepAnT~\cite{munir2018deepant} & 0\% & 0\% & 89\% (\checkmark) \\
RobustPCA~\cite{paffenroth2018robust} & 0\% & 0\% & 100\% (\checkmark) \\
Hybrid $k$NN~\cite{song2017hybrid} & 0\% & 0\% & 98\% (\checkmark) \\ \shline
\end{tabular}
}
\end{center}
\end{table} 

%% file: insert/tab_benchmark_1.tex
\begin{table*}[htp]
\caption{
Overview of performance for all algorithms, with box plots illustrating the distribution of AUC-ROC and AUC-P$_\text{T}$R$_\text{T}$ for unsupervised, supervised, and semi-supervised learning datasets. 
The orange lines represent medians, while the green lines indicate means. 
The Matrix Profile-based anomaly detection system consistently delivers high performance across all setups.
}
\label{tab:benchmark}
\begin{center}
\resizebox{1.0\linewidth}{!}{%
\begin{tabular}{l||cc|cc|cc}
\shline
\multirow{2}{*}{Method} & \multicolumn{2}{c|}{Unsupervised} & \multicolumn{2}{c|}{Supervised} & \multicolumn{2}{c}{Semi-supervised} \\
& AUC-ROC & AUC-P$_\text{T}$R$_\text{T}$ & AUC-ROC & AUC-P$_\text{T}$R$_\text{T}$ & AUC-ROC & AUC-P$_\text{T}$R$_\text{T}$ \\ \hline \hline
Matrix Profile & \begin{minipage}{0.15\linewidth}\includegraphics[width=\linewidth]{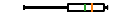}\end{minipage} & \begin{minipage}{0.15\linewidth}\includegraphics[width=\linewidth]{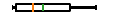}\end{minipage} & \begin{minipage}{0.15\linewidth}\includegraphics[width=\linewidth]{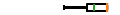}\end{minipage} & \begin{minipage}{0.15\linewidth}\includegraphics[width=\linewidth]{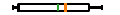}\end{minipage} & \begin{minipage}{0.15\linewidth}\includegraphics[width=\linewidth]{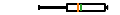}\end{minipage} & \begin{minipage}{0.15\linewidth}\includegraphics[width=\linewidth]{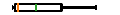}\end{minipage} \\
$k$-Means~\cite{yairi2001fault} & \begin{minipage}{0.15\linewidth}\includegraphics[width=\linewidth]{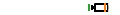}\end{minipage} & \begin{minipage}{0.15\linewidth}\includegraphics[width=\linewidth]{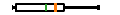}\end{minipage} & \textcolor{lightgray}{not available} & \textcolor{lightgray}{not available} & \textcolor{lightgray}{not available} & \textcolor{lightgray}{not available} \\
EIF~\cite{hariri2019extended} & \begin{minipage}{0.15\linewidth}\includegraphics[width=\linewidth]{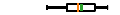}\end{minipage} & \begin{minipage}{0.15\linewidth}\includegraphics[width=\linewidth]{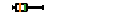}\end{minipage} & \textcolor{lightgray}{not available} & \textcolor{lightgray}{not available} & \begin{minipage}{0.15\linewidth}\includegraphics[width=\linewidth]{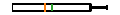}\end{minipage} & \begin{minipage}{0.15\linewidth}\includegraphics[width=\linewidth]{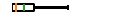}\end{minipage} \\
iForest~\cite{liu2008isolation} & \begin{minipage}{0.15\linewidth}\includegraphics[width=\linewidth]{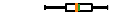}\end{minipage} & \begin{minipage}{0.15\linewidth}\includegraphics[width=\linewidth]{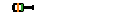}\end{minipage} & \textcolor{lightgray}{not available} & \textcolor{lightgray}{not available} & \begin{minipage}{0.15\linewidth}\includegraphics[width=\linewidth]{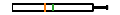}\end{minipage} & \begin{minipage}{0.15\linewidth}\includegraphics[width=\linewidth]{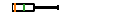}\end{minipage} \\
$k$NN~\cite{ramaswamy2000efficient} & \begin{minipage}{0.15\linewidth}\includegraphics[width=\linewidth]{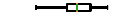}\end{minipage} & \begin{minipage}{0.15\linewidth}\includegraphics[width=\linewidth]{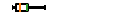}\end{minipage} & \textcolor{lightgray}{not available} & \textcolor{lightgray}{not available} & \begin{minipage}{0.15\linewidth}\includegraphics[width=\linewidth]{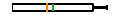}\end{minipage} & \begin{minipage}{0.15\linewidth}\includegraphics[width=\linewidth]{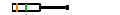}\end{minipage} \\
CBLOF~\cite{he2003discovering} & \begin{minipage}{0.15\linewidth}\includegraphics[width=\linewidth]{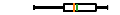}\end{minipage} & \begin{minipage}{0.15\linewidth}\includegraphics[width=\linewidth]{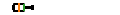}\end{minipage} & \textcolor{lightgray}{not available} & \textcolor{lightgray}{not available} & \begin{minipage}{0.15\linewidth}\includegraphics[width=\linewidth]{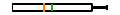}\end{minipage} & \begin{minipage}{0.15\linewidth}\includegraphics[width=\linewidth]{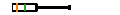}\end{minipage} \\
IF-LOF~\cite{cheng2019outlier} & \begin{minipage}{0.15\linewidth}\includegraphics[width=\linewidth]{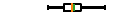}\end{minipage} & \begin{minipage}{0.15\linewidth}\includegraphics[width=\linewidth]{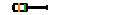}\end{minipage} & \textcolor{lightgray}{not available} & \textcolor{lightgray}{not available} & \textcolor{lightgray}{not available} & \textcolor{lightgray}{not available} \\
LOF~\cite{breunig2000lof} & \begin{minipage}{0.15\linewidth}\includegraphics[width=\linewidth]{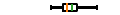}\end{minipage} & \begin{minipage}{0.15\linewidth}\includegraphics[width=\linewidth]{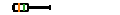}\end{minipage} & \textcolor{lightgray}{not available} & \textcolor{lightgray}{not available} & \begin{minipage}{0.15\linewidth}\includegraphics[width=\linewidth]{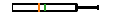}\end{minipage} & \begin{minipage}{0.15\linewidth}\includegraphics[width=\linewidth]{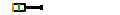}\end{minipage} \\
COPOD~\cite{li2020copod} & \begin{minipage}{0.15\linewidth}\includegraphics[width=\linewidth]{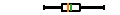}\end{minipage} & \begin{minipage}{0.15\linewidth}\includegraphics[width=\linewidth]{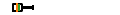}\end{minipage} & \textcolor{lightgray}{not available} & \textcolor{lightgray}{not available} & \begin{minipage}{0.15\linewidth}\includegraphics[width=\linewidth]{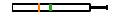}\end{minipage} & \begin{minipage}{0.15\linewidth}\includegraphics[width=\linewidth]{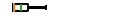}\end{minipage} \\
PCC~\cite{shyu2003novel} & \begin{minipage}{0.15\linewidth}\includegraphics[width=\linewidth]{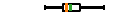}\end{minipage} & \begin{minipage}{0.15\linewidth}\includegraphics[width=\linewidth]{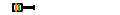}\end{minipage} & \textcolor{lightgray}{not available} & \textcolor{lightgray}{not available} & \begin{minipage}{0.15\linewidth}\includegraphics[width=\linewidth]{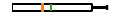}\end{minipage} & \begin{minipage}{0.15\linewidth}\includegraphics[width=\linewidth]{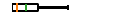}\end{minipage} \\
HBOS~\cite{goldstein2012histogram} & \begin{minipage}{0.15\linewidth}\includegraphics[width=\linewidth]{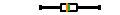}\end{minipage} & \begin{minipage}{0.15\linewidth}\includegraphics[width=\linewidth]{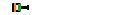}\end{minipage} & \textcolor{lightgray}{not available} & \textcolor{lightgray}{not available} & \begin{minipage}{0.15\linewidth}\includegraphics[width=\linewidth]{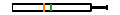}\end{minipage} & \begin{minipage}{0.15\linewidth}\includegraphics[width=\linewidth]{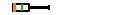}\end{minipage} \\
Torsk~\cite{heim2019adaptive} & \begin{minipage}{0.15\linewidth}\includegraphics[width=\linewidth]{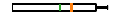}\end{minipage} & \begin{minipage}{0.15\linewidth}\includegraphics[width=\linewidth]{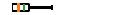}\end{minipage} & \textcolor{lightgray}{not available} & \textcolor{lightgray}{not available} & \textcolor{lightgray}{not available} & \textcolor{lightgray}{not available} \\
COF~\cite{tang2002enhancing} & \begin{minipage}{0.15\linewidth}\includegraphics[width=\linewidth]{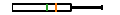}\end{minipage} & \begin{minipage}{0.15\linewidth}\includegraphics[width=\linewidth]{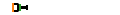}\end{minipage} & \textcolor{lightgray}{not available} & \textcolor{lightgray}{not available} & \textcolor{lightgray}{not available} & \textcolor{lightgray}{not available} \\
HIF~\cite{marteau2017hybrid} & \textcolor{lightgray}{not available} & \textcolor{lightgray}{not available} & \begin{minipage}{0.15\linewidth}\includegraphics[width=\linewidth]{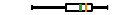}\end{minipage} & \begin{minipage}{0.15\linewidth}\includegraphics[width=\linewidth]{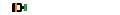}\end{minipage} & \textcolor{lightgray}{not available} & \textcolor{lightgray}{not available} \\
MultiHMM~\cite{li2017multivariate} & \textcolor{lightgray}{not available} & \textcolor{lightgray}{not available} & \begin{minipage}{0.15\linewidth}\includegraphics[width=\linewidth]{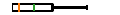}\end{minipage} & \begin{minipage}{0.15\linewidth}\includegraphics[width=\linewidth]{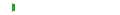}\end{minipage} & \textcolor{lightgray}{not available} & \textcolor{lightgray}{not available} \\
Telemanom~\cite{hundman2018detecting} & \textcolor{lightgray}{not available} & \textcolor{lightgray}{not available} & \textcolor{lightgray}{not available} & \textcolor{lightgray}{not available} & \begin{minipage}{0.15\linewidth}\includegraphics[width=\linewidth]{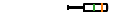}\end{minipage} & \begin{minipage}{0.15\linewidth}\includegraphics[width=\linewidth]{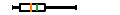}\end{minipage} \\
OmniAnomaly~\cite{su2019robust} & \textcolor{lightgray}{not available} & \textcolor{lightgray}{not available} & \textcolor{lightgray}{not available} & \textcolor{lightgray}{not available} & \begin{minipage}{0.15\linewidth}\includegraphics[width=\linewidth]{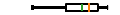}\end{minipage} & \begin{minipage}{0.15\linewidth}\includegraphics[width=\linewidth]{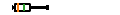}\end{minipage} \\
DeepAnT~\cite{munir2018deepant} & \textcolor{lightgray}{not available} & \textcolor{lightgray}{not available} & \textcolor{lightgray}{not available} & \textcolor{lightgray}{not available} & \begin{minipage}{0.15\linewidth}\includegraphics[width=\linewidth]{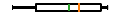}\end{minipage} & \begin{minipage}{0.15\linewidth}\includegraphics[width=\linewidth]{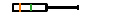}\end{minipage} \\
RobustPCA~\cite{paffenroth2018robust} & \textcolor{lightgray}{not available} & \textcolor{lightgray}{not available} & \textcolor{lightgray}{not available} & \textcolor{lightgray}{not available} & \begin{minipage}{0.15\linewidth}\includegraphics[width=\linewidth]{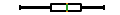}\end{minipage} & \begin{minipage}{0.15\linewidth}\includegraphics[width=\linewidth]{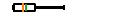}\end{minipage} \\
Hybrid $k$NN~\cite{song2017hybrid} & \textcolor{lightgray}{not available} \vspace{-0.1em} & \textcolor{lightgray}{not available} \vspace{-0.1em} & \textcolor{lightgray}{not available} \vspace{-0.1em} & \textcolor{lightgray}{not available} \vspace{-0.1em} & \begin{minipage}{0.15\linewidth}\includegraphics[width=\linewidth]{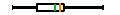}\end{minipage} \vspace{-0.1em} & \begin{minipage}{0.15\linewidth}\includegraphics[width=\linewidth]{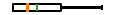}\end{minipage} \vspace{-0.1em} \\
& \begin{minipage}{0.15\linewidth}\includegraphics[width=\linewidth]{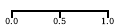}\end{minipage} & \begin{minipage}{0.15\linewidth}\includegraphics[width=\linewidth]{insert/figure/benchmark_1/xaxis.pdf}\end{minipage} & \begin{minipage}{0.15\linewidth}\includegraphics[width=\linewidth]{insert/figure/benchmark_1/xaxis.pdf}\end{minipage} & \begin{minipage}{0.15\linewidth}\includegraphics[width=\linewidth]{insert/figure/benchmark_1/xaxis.pdf}\end{minipage} & \begin{minipage}{0.15\linewidth}\includegraphics[width=\linewidth]{insert/figure/benchmark_1/xaxis.pdf}\end{minipage} & \begin{minipage}{0.15\linewidth}\includegraphics[width=\linewidth]{insert/figure/benchmark_1/xaxis.pdf}\end{minipage} \\ \shline
\end{tabular}
}
\end{center}
\end{table*} 

%% file: insert/fig_ad_runtime.tex
\begin{figure}[htp]
\centerline{
\includegraphics[width=0.99\linewidth]{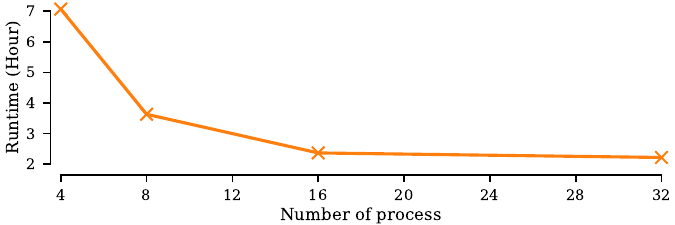}
}
\caption{
Total benchmark runtime for various number of processes settings.
}
\label{fig:ad_runtime}
\end{figure}

%% file: insert/fig_param_sub.tex
\begin{figure}[htp]
\centerline{
\includegraphics[width=0.99\linewidth]{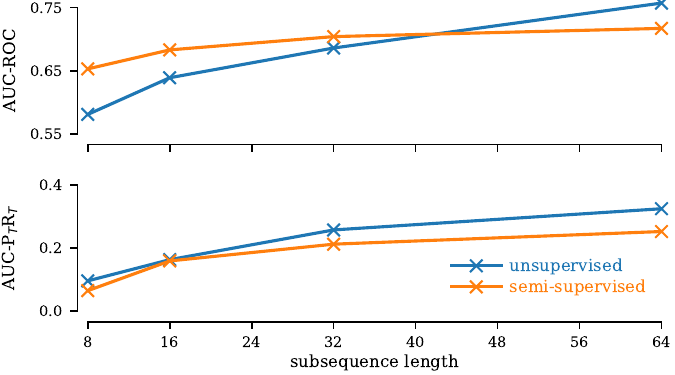}
}
\caption{
The relationship between anomaly detection performance and subsequence length for unsupervised and semi-supervised learning datasets. 
The subsequence length parameter is crucial and can greatly impact the performance of anomaly detection.}
\label{fig:param_sub}
\end{figure}

%% file: insert/fig_param_knn.tex
\begin{figure}[htp]
\centerline{
\includegraphics[width=0.99\linewidth]{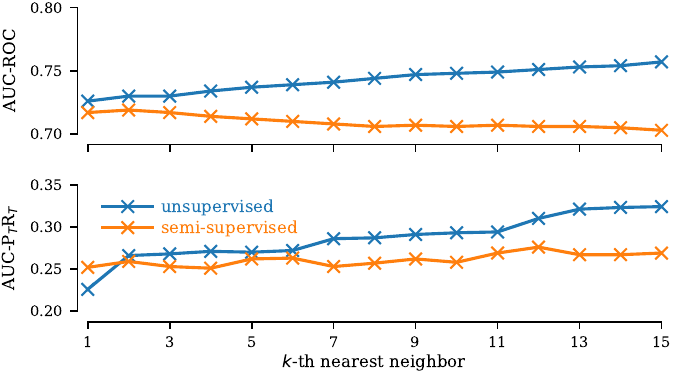}
}
\caption{
The relationship between anomaly detection performance and the parameter $k$ for finding the $k$-th nearest neighbor in unsupervised and semi-supervised learning datasets. 
While $k$ has a minimal impact on the performance of semi-supervised learning datasets, increasing the value of $k$ enhances performance for unsupervised learning datasets.
}
\label{fig:param_knn}
\end{figure}

%% file: insert/tab_param_1.tex
\begin{table}[htp]
\caption{Performance of the Matrix Profile-based anomaly detection system with different hyper-parameter settings for both unsupervised and semi-supervised learning.}
\label{tab:param}
\begin{center}
\resizebox{\linewidth}{!}{%
\begin{tabular}{ll||cc|cc}
\shline
                                                                               &       & \multicolumn{2}{c|}{Unsupervised} & \multicolumn{2}{c}{Semi-supervised}                        \\
                                                                               &       & AUC-ROC         & AUC-PTRT       & \multicolumn{1}{c}{AUC-ROC} & \multicolumn{1}{c}{AUC-PTRT} \\ \hline \hline
\multirow{2}{*}{\begin{tabular}[c]{@{}l@{}}Sorting\\ Placement\end{tabular}}   & Pre   & 0.757$\pm$0.240    & 0.324$\pm$0.321   & 0.717$\pm$0.226                & 0.252$\pm$0.334                 \\
                                                                               & Post  & 0.760$\pm$0.220    & 0.275$\pm$0.321   & 0.682$\pm$0.271                & 0.240$\pm$0.323                 \\ \hline
\multirow{2}{*}{\begin{tabular}[c]{@{}l@{}}Dimension\\ Selection\end{tabular}} & First & 0.757$\pm$0.240    & 0.324$\pm$0.321   & 0.717$\pm$0.226                & 0.252$\pm$0.334                 \\
                                                                               & Mean  & 0.757$\pm$0.222    & 0.250$\pm$0.288   & 0.692$\pm$0.254                & 0.234$\pm$0.317                 \\ \hline
\multirow{2}{*}{Post-process}                                                   & Yes   & 0.757$\pm$0.240    & 0.324$\pm$0.321   & 0.717$\pm$0.226                & 0.252$\pm$0.334                 \\
                                                                               & No    & 0.752$\pm$0.220    & 0.291$\pm$0.264   & 0.695$\pm$0.200                & 0.202$\pm$0.271\\ \shline
\end{tabular}
}
\end{center}
\end{table} 

%% file: section/conclusion.tex
\section{Conclusion}
In this paper, we have provided a comprehensive study on applying the Matrix Profile (MP) to a multidimensional time series anomaly detection (TSAD) problem. 
We compared different methods for generating multidimensional MPs from pairwise distance tensors, proposed an efficient $k$-nearest neighbor MP extension, and discussed how to utilize various types of joins offered by the MP for TSAD. 
Using the knowledge presented, we designed an MP-based TSAD system with competitive performance on 119 datasets, compared to the other 19 baseline methods. 
Beyond the algorithmic improvements presented in this work, we provide a complete open-source implementation of our method. 
Within this repository, we have also included newly expanded benchmark results evaluating our approach against the rigorous TSB-AD framework \cite{yeh2026matrix, yeh2026mmpad}, providing a robust foundation for future comparative studies. 
For future work, we would like to explore the interaction between MP-based multidimensional TSAD methods with sketching~\cite{yeh2023sketching}, random projection~\cite{yeh2024rpmixer}, or dictionary learning~\cite{yeh2022error}. 
It would also be interesting to explore the possibility of combining pretrained/foundation models~\cite{yeh2023toward,der2024systematic} with an MP-based multidimensional TSAD system.

%% file: section/reference.bib
@misc{supplementary,
  title={Project Website},
  author={{The Author(s)}},
  note = {\url{https://sites.google.com/view/mp4ad}},
  year={2024}
}

@inproceedings{yeh2016matrix,
  title={Matrix profile I: all pairs similarity joins for time series: a unifying view that includes motifs, discords and shapelets},
  author={Yeh, Chin-Chia Michael and Zhu, Yan and Ulanova, Liudmila and Begum, Nurjahan and Ding, Yifei and Dau, Hoang Anh and Silva, Diego Furtado and Mueen, Abdullah and Keogh, Eamonn},
  booktitle={2016 IEEE 16th international conference on data mining (ICDM)},
  pages={1317--1322},
  year={2016},
  organization={Ieee}
}

@inproceedings{yeh2017matrix,
  title={Matrix profile VI: Meaningful multidimensional motif discovery},
  author={Yeh, Chin-Chia Michael and Kavantzas, Nickolas and Keogh, Eamonn},
  booktitle={2017 IEEE international conference on data mining (ICDM)},
  pages={565--574},
  year={2017},
  organization={IEEE}
}

@inproceedings{yeh2022error,
  title={Error-bounded approximate time series joins using compact dictionary representations of time series},
  author={Yeh, Chin-Chia Michael and Zheng, Yan and Wang, Junpeng and Chen, Huiyuan and Zhuang, Zhongfang and Zhang, Wei and Keogh, Eamonn},
  booktitle={Proceedings of the 2022 SIAM International Conference on Data Mining (SDM)},
  pages={181--189},
  year={2022},
  organization={SIAM}
}

@article{yeh2023sketching,
  title={Sketching multidimensional time series for fast discord mining},
  author={Yeh, Chin-Chia Michael and Zheng, Yan and Pan, Menghai and Chen, Huiyuan and Zhuang, Zhongfang and Wang, Junpeng and Wang, Liang and Zhang, Wei and Phillips, Jeff M and Keogh, Eamonn},
  journal={arXiv preprint arXiv:2311.03393},
  year={2023}
}

@book{yeh2018towards,
  title={Towards a near universal time series data mining tool: Introducing the matrix profile},
  author={Yeh, Michael Chin-Chia},
  year={2018},
  publisher={University of California, Riverside}
}

@inproceedings{yeh2023toward,
  title={Toward a foundation model for time series data},
  author={Yeh, Chin-Chia Michael and Dai, Xin and Chen, Huiyuan and Zheng, Yan and Fan, Yujie and Der, Audrey and Lai, Vivian and Zhuang, Zhongfang and Wang, Junpeng and Wang, Liang and others},
  booktitle={Proceedings of the 32nd ACM International Conference on Information and Knowledge Management},
  pages={4400--4404},
  year={2023}
}

@article{schmidl2022anomaly,
  title={Anomaly detection in time series: a comprehensive evaluation},
  author={Schmidl, Sebastian and Wenig, Phillip and Papenbrock, Thorsten},
  journal={Proceedings of the VLDB Endowment},
  volume={15},
  number={9},
  pages={1779--1797},
  year={2022},
  publisher={VLDB Endowment}
}

@article{bradley1997use,
  title={The use of the area under the ROC curve in the evaluation of machine learning algorithms},
  author={Bradley, Andrew P},
  journal={Pattern recognition},
  volume={30},
  number={7},
  pages={1145--1159},
  year={1997},
  publisher={Elsevier}
}

@article{hanley1982meaning,
  title={The meaning and use of the area under a receiver operating characteristic (ROC) curve.},
  author={Hanley, James A and McNeil, Barbara J},
  journal={Radiology},
  volume={143},
  number={1},
  pages={29--36},
  year={1982}
}

@article{tatbul2018precision,
  title={Precision and recall for time series},
  author={Tatbul, Nesime and Lee, Tae Jun and Zdonik, Stan and Alam, Mejbah and Gottschlich, Justin},
  journal={Advances in neural information processing systems},
  volume={31},
  year={2018}
}

@inproceedings{tafazoli2023matrix,
  title={Matrix Profile XXVIII: Discovering Multi-Dimensional Time Series Anomalies with K of N Anomaly Detection},
  author={Tafazoli, Sadaf and Keogh, Eamonn},
  booktitle={Proceedings of the 2023 SIAM International Conference on Data Mining (SDM)},
  pages={685--693},
  year={2023},
  organization={SIAM}
}

@inproceedings{lu2022matrix,
  title={Matrix profile XXIV: scaling time series anomaly detection to trillions of datapoints and ultra-fast arriving data streams},
  author={Lu, Yue and Wu, Renjie and Mueen, Abdullah and Zuluaga, Maria A and Keogh, Eamonn},
  booktitle={Proceedings of the 28th ACM SIGKDD Conference on Knowledge Discovery and Data Mining},
  pages={1173--1182},
  year={2022}
}

@inproceedings{keogh2005hot,
  title={Hot sax: Efficiently finding the most unusual time series subsequence},
  author={Keogh, Eamonn and Lin, Jessica and Fu, Ada},
  booktitle={Fifth IEEE International Conference on Data Mining (ICDM'05)},
  pages={8--pp},
  year={2005},
  organization={Ieee}
}

@inproceedings{zhu2016matrix,
  title={Matrix profile ii: Exploiting a novel algorithm and gpus to break the one hundred million barrier for time series motifs and joins},
  author={Zhu, Yan and Zimmerman, Zachary and Senobari, Nader Shakibay and Yeh, Chin-Chia Michael and Funning, Gareth and Mueen, Abdullah and Brisk, Philip and Keogh, Eamonn},
  booktitle={2016 IEEE 16th international conference on data mining (ICDM)},
  pages={739--748},
  year={2016},
  organization={IEEE}
}

@article{hoare1961algorithm,
  title={Algorithm 65: find},
  author={Hoare, Charles AR},
  journal={Communications of the ACM},
  volume={4},
  number={7},
  pages={321--322},
  year={1961},
  publisher={ACM New York, NY, USA}
}

@inproceedings{lu2023matrix,
  title={Matrix Profile XXX: MADRID: A Hyper-Anytime and Parameter-Free Algorithm to Find Time Series Anomalies of all Lengths},
  author={Lu, Yue and Srinivas, Thirumalai Vinjamoor Akhil and Nakamura, Takaaki and Imamura, Makoto and Keogh, Eamonn},
  booktitle={2023 IEEE International Conference on Data Mining (ICDM)},
  pages={1199--1204},
  year={2023},
  organization={IEEE}
}

@article{wenig2022timeeval,
  title={TimeEval: A benchmarking toolkit for time series anomaly detection algorithms},
  author={Wenig, Phillip and Schmidl, Sebastian and Papenbrock, Thorsten},
  journal={Proceedings of the VLDB Endowment},
  volume={15},
  number={12},
  pages={3678--3681},
  year={2022},
  publisher={VLDB Endowment}
}

@misc{hutchins2006calit2,
  author={Hutchins,Jon},
  title={{CalIt2 Building People Counts}},
  year={2006},
  howpublished={UCI Machine Learning Repository},
  note={{DOI}: https://doi.org/10.24432/C5NG78}
}

@article{bachlin2009wearable,
  title={Wearable assistant for Parkinson’s disease patients with the freezing of gait symptom},
  author={Bachlin, Marc and Plotnik, Meir and Roggen, Daniel and Maidan, Inbal and Hausdorff, Jeffrey M and Giladi, Nir and Troster, Gerhard},
  journal={IEEE Transactions on Information Technology in Biomedicine},
  volume={14},
  number={2},
  pages={436--446},
  year={2009},
  publisher={IEEE}
}

@article{jacob2020exathlon,
  title={Exathlon: A benchmark for explainable anomaly detection over time series},
  author={Jacob, Vincent and Song, Fei and Stiegler, Arnaud and Rad, Bijan and Diao, Yanlei and Tatbul, Nesime},
  journal={arXiv preprint arXiv:2010.05073},
  year={2020}
}

@article{von2018anomaly,
  title={Anomaly detection and localization for cyber-physical production systems with self-organizing maps},
  author={von Birgelen, Alexander and Niggemann, Oliver},
  journal={IMPROVE-Innovative Modelling Approaches for Production Systems to Raise Validatable Efficiency: Intelligent Methods for the Factory of the Future},
  pages={55--71},
  year={2018},
  publisher={Springer Berlin Heidelberg}
}

@article{goldberger2000physiobank,
  title={PhysioBank, PhysioToolkit, and PhysioNet: components of a new research resource for complex physiologic signals},
  author={Goldberger, Ary L and Amaral, Luis AN and Glass, Leon and Hausdorff, Jeffrey M and Ivanov, Plamen Ch and Mark, Roger G and Mietus, Joseph E and Moody, George B and Peng, Chung-Kang and Stanley, H Eugene},
  journal={circulation},
  volume={101},
  number={23},
  pages={e215--e220},
  year={2000},
  publisher={Am Heart Assoc}
}

@inproceedings{su2019robust,
  title={Robust anomaly detection for multivariate time series through stochastic recurrent neural network},
  author={Su, Ya and Zhao, Youjian and Niu, Chenhao and Liu, Rong and Sun, Wei and Pei, Dan},
  booktitle={Proceedings of the 25th ACM SIGKDD international conference on knowledge discovery \& data mining},
  pages={2828--2837},
  year={2019}
}

@article{musser1997introspective,
  title={Introspective sorting and selection algorithms},
  author={Musser, David R},
  journal={Software: Practice and Experience},
  volume={27},
  number={8},
  pages={983--993},
  year={1997},
  publisher={Wiley Online Library}
}

@inproceedings{shyu2003novel,
  title={A novel anomaly detection scheme based on principal component classifier},
  author={Shyu, Mei-Ling and Chen, Shu-Ching and Sarinnapakorn, Kanoksri and Chang, LiWu},
  booktitle={Proceedings of the IEEE foundations and new directions of data mining workshop},
  pages={172--179},
  year={2003},
  organization={IEEE Press}
}

@article{hariri2019extended,
  title={Extended isolation forest},
  author={Hariri, Sahand and Kind, Matias Carrasco and Brunner, Robert J},
  journal={IEEE transactions on knowledge and data engineering},
  volume={33},
  number={4},
  pages={1479--1489},
  year={2019},
  publisher={IEEE}
}

@inproceedings{yairi2001fault,
  title={Fault detection by mining association rules from house-keeping data},
  author={Yairi, Takehisa and Kato, Yoshikiyo and Hori, Koichi},
  booktitle={proceedings of the 6th International Symposium on Artificial Intelligence, Robotics and Automation in Space},
  volume={18},
  pages={21},
  year={2001},
  organization={Citeseer}
}

@inproceedings{liu2008isolation,
  title={Isolation forest},
  author={Liu, Fei Tony and Ting, Kai Ming and Zhou, Zhi-Hua},
  booktitle={2008 eighth ieee international conference on data mining},
  pages={413--422},
  year={2008},
  organization={IEEE}
}

@inproceedings{ramaswamy2000efficient,
  title={Efficient algorithms for mining outliers from large data sets},
  author={Ramaswamy, Sridhar and Rastogi, Rajeev and Shim, Kyuseok},
  booktitle={Proceedings of the 2000 ACM SIGMOD international conference on Management of data},
  pages={427--438},
  year={2000}
}

@article{heim2019adaptive,
  title={Adaptive anomaly detection in chaotic time series with a spatially aware echo state network},
  author={Heim, Niklas and Avery, James E},
  journal={arXiv preprint arXiv:1909.01709},
  year={2019}
}

@article{he2003discovering,
  title={Discovering cluster-based local outliers},
  author={He, Zengyou and Xu, Xiaofei and Deng, Shengchun},
  journal={Pattern recognition letters},
  volume={24},
  number={9-10},
  pages={1641--1650},
  year={2003},
  publisher={Elsevier}
}

@inproceedings{tang2002enhancing,
  title={Enhancing effectiveness of outlier detections for low density patterns},
  author={Tang, Jian and Chen, Zhixiang and Fu, Ada Wai-Chee and Cheung, David W},
  booktitle={Advances in Knowledge Discovery and Data Mining: 6th Pacific-Asia Conference, PAKDD 2002 Taipei, Taiwan, May 6--8, 2002 Proceedings 6},
  pages={535--548},
  year={2002},
  organization={Springer}
}

@inproceedings{cheng2019outlier,
  title={Outlier detection using isolation forest and local outlier factor},
  author={Cheng, Zhangyu and Zou, Chengming and Dong, Jianwei},
  booktitle={Proceedings of the conference on research in adaptive and convergent systems},
  pages={161--168},
  year={2019}
}

@inproceedings{breunig2000lof,
  title={LOF: identifying density-based local outliers},
  author={Breunig, Markus M and Kriegel, Hans-Peter and Ng, Raymond T and Sander, J{\"o}rg},
  booktitle={Proceedings of the 2000 ACM SIGMOD international conference on Management of data},
  pages={93--104},
  year={2000}
}

@inproceedings{li2020copod,
  title={COPOD: copula-based outlier detection},
  author={Li, Zheng and Zhao, Yue and Botta, Nicola and Ionescu, Cezar and Hu, Xiyang},
  booktitle={2020 IEEE international conference on data mining (ICDM)},
  pages={1118--1123},
  year={2020},
  organization={IEEE}
}

@article{goldstein2012histogram,
  title={Histogram-based outlier score (hbos): A fast unsupervised anomaly detection algorithm},
  author={Goldstein, Markus and Dengel, Andreas},
  journal={KI-2012: poster and demo track},
  volume={1},
  pages={59--63},
  year={2012},
  publisher={Citeseer}
}

@article{marteau2017hybrid,
  title={Hybrid isolation forest-application to intrusion detection},
  author={Marteau, Pierre-Fran{\c{c}}ois and Soheily-Khah, Saeid and B{\'e}chet, Nicolas},
  journal={arXiv preprint arXiv:1705.03800},
  year={2017}
}

@article{li2017multivariate,
  title={Multivariate time series anomaly detection: A framework of Hidden Markov Models},
  author={Li, Jinbo and Pedrycz, Witold and Jamal, Iqbal},
  journal={Applied Soft Computing},
  volume={60},
  pages={229--240},
  year={2017},
  publisher={Elsevier}
}

@article{song2017hybrid,
  title={A hybrid semi-supervised anomaly detection model for high-dimensional data},
  author={Song, Hongchao and Jiang, Zhuqing and Men, Aidong and Yang, Bo and others},
  journal={Computational intelligence and neuroscience},
  volume={2017},
  year={2017},
  publisher={Hindawi}
}

@inproceedings{hundman2018detecting,
  title={Detecting spacecraft anomalies using lstms and nonparametric dynamic thresholding},
  author={Hundman, Kyle and Constantinou, Valentino and Laporte, Christopher and Colwell, Ian and Soderstrom, Tom},
  booktitle={Proceedings of the 24th ACM SIGKDD international conference on knowledge discovery \& data mining},
  pages={387--395},
  year={2018}
}

@article{munir2018deepant,
  title={DeepAnT: A deep learning approach for unsupervised anomaly detection in time series},
  author={Munir, Mohsin and Siddiqui, Shoaib Ahmed and Dengel, Andreas and Ahmed, Sheraz},
  journal={Ieee Access},
  volume={7},
  pages={1991--2005},
  year={2018},
  publisher={IEEE}
}

@article{paffenroth2018robust,
  title={Robust pca for anomaly detection in cyber networks},
  author={Paffenroth, Randy and Kay, Kathleen and Servi, Les},
  journal={arXiv preprint arXiv:1801.01571},
  year={2018}
}

@inproceedings{baumgartner2023mtads,
  title={mTADS: Multivariate Time Series Anomaly Detection Benchmark Suites},
  author={Baumgartner, David and Langseth, Helge and Ramampiaro, Heri and Eng{\o}-Monsen, Kenth},
  booktitle={2023 IEEE International Conference on Big Data (BigData)},
  pages={588--597},
  year={2023},
  organization={IEEE}
}

@inproceedings{yeh2024rpmixer,
  title={RPMixer: Shaking Up Time Series Forecasting with Random Projections for Large Spatial-Temporal Data},
  author={Yeh, Chin-Chia Michael and Fan, Yujie and Dai, Xin and Saini, Uday Singh and Lai, Vivian and Aboagye, Prince Osei and Wang, Junpeng and Chen, Huiyuan and Zheng, Yan and Zhuang, Zhongfang and others},
  booktitle={Proceedings of the 30th ACM SIGKDD Conference on Knowledge Discovery and Data Mining},
  pages={3919--3930},
  year={2024}
}

@article{der2024systematic,
  title={A Systematic Evaluation of Generated Time Series and Their Effects in Self-Supervised Pretraining},
  author={Der, Audrey and Yeh, Chin-Chia Michael and Dai, Xin and Chen, Huiyuan and Zheng, Yan and Fan, Yujie and Zhuang, Zhongfang and Lai, Vivian and Wang, Junpeng and Wang, Liang and others},
  journal={arXiv preprint arXiv:2408.07869},
  year={2024}
}

@misc{yeh2026matrix,
  title={Matrix Profile for Time-Series Anomaly Detection: A Reproducible Open-Source Benchmark on TSB-AD}, 
  author={Yeh, Chin-Chia Michael},
  year={2026},
  eprint={2604.02445},
  archivePrefix={arXiv},
  url={https://arxiv.org/abs/2604.02445}
}

@article{liu2024elephant,
  title={The elephant in the room: Towards a reliable time-series anomaly detection benchmark},
  author={Liu, Qinghua and Paparrizos, John},
  journal={Advances in Neural Information Processing Systems},
  volume={37},
  pages={108231--108261},
  year={2024}
}

@misc{yeh2026mmpad,
  author = {Yeh, Chin-Chia Michael},
  title = {MMPAD submission for TSB-AD: Code and archived results.},
  url = {https://github.com/mcyeh/mmpad_tsb},
  year = {2026}
}
